\documentclass[10pt,twocolumn,letterpaper]{article}

\usepackage{cvpr}              

\usepackage[dvipsnames]{xcolor}

\definecolor{cvprblue}{rgb}{0.21,0.49,0.74}
\usepackage[pagebackref,breaklinks,colorlinks,citecolor=cvprblue]{hyperref}

\def\paperID{*****} 
\def\confName{CVPR}
\def\confYear{2024}

\title{The 6th Affective Behavior Analysis in-the-wild (ABAW) Competition}

\author{Dimitrios Kollias \\
Queen Mary University of London, UK\\
{\tt\small d.kollias@qmul.ac.uk}
\and
Panagiotis Tzirakis \\
Hume AI, USA \\
{\tt\small panagiotis@hume.ai}
\and
Alan Cowen\\
Hume AI, USA \\
\and
Stefanos Zafeiriou\\
Imperial College London, UK \\
\and
Irene Kotsia\\
Cogitat, UK \\
\and
\and
Alice Baird\\
Hume AI, USA \\
\and
Chris Gagne\\
Hume AI, USA \\
\and
Chunchang Shao \\
Queen Mary University of London, UK\\
\and
Guanyu Hu \\
Queen Mary University of London, UK\\
Xi'an Jiaotong University, China \\
}

\begin{document}
\maketitle
\begin{abstract}
This paper describes the 6th Affective Behavior Analysis in-the-wild (ABAW)  Competition, which is part of the respective Workshop held in conjunction with IEEE CVPR 2024. The 6th ABAW Competition addresses contemporary challenges in understanding human emotions and behaviors, crucial for the development of human-centered technologies. In more detail, 
 the Competition focuses on affect related benchmarking
tasks and comprises of five sub-challenges: i) Valence-Arousal Estimation (the target is to estimate two continuous affect dimensions, valence and arousal), ii)  Expression Recognition (the target is to recognise between the mutually exclusive classes of the 7 basic expressions and 'other'), iii)  Action Unit Detection  (the target is to detect 12 action units), iv) Compound Expression Recognition (the target is to recognise between the 7 mutually exclusive compound expression classes), and v) Emotional Mimicry Intensity Estimation (the target is to estimate six continuous emotion dimensions).
In the paper, we present these Challenges, describe their respective datasets and challenge protocols (we outline the evaluation metrics) and present  the baseline systems as well as their obtained performance. More information for the Competition can be found in: \url{https://affective-behavior-analysis-in-the-wild.github.io/6th}.
\end{abstract}    
\section{Introduction}
\label{sec:intro}

The 6th Affective Behavior Analysis in-the-wild (ABAW) Workshop and Competition continues its tradition of fostering interdisciplinary collaboration by bringing together experts from various fields including academia, industry, and government. This workshop, held in conjunction with IEEE Computer Vision and Pattern Recognition Conference (CVPR) 2024, aims to delve into the analysis of affective behavior in real-world settings, a critical aspect for the development of human-centered technologies such as HCI systems and intelligent digital assistants. By understanding human emotions and behaviors, machines can better engage with users irrespective of contextual factors like age, gender, or social background, thereby enhancing trust and interaction in real-life scenarios.

The ABAW Competition, an integral part of the workshop, is split into five Challenges. 

The first Challenge is the Valence-Arousal (VA) Estimation one; the target of this Challenge is to estimate the two continuous affect dimensions of valence and
arousal in each frame of the utilized Challenge corpora. Valence  characterises an affective state on a continuous scale from positive to negative (in other words from -1 to 1). Arousal characterises an affective state on a continuous scale from active to passive (in other words from -1 to 1).

Only uni-task solutions will be accepted for this Challenge. Teams are allowed to use any -publicly or not- available pre-trained model (as long as it has not been pre-trained on the utilized in this Challenge corpora, Aff-Wild2). The pre-trained model can be pre-trained on any task (e.g., VA estimation, Expression Recognition, AU detection, Face Recognition). However when the teams are refining the model and developing the methodology they should not use any other annotations (expressions or AUs): the methodology should be purely uni-task, using only the VA annotations. This means that teams are allowed to use other databases' VA annotations, or generated/synthetic data, or any affine transformations, or in general data augmentation techniques for increasing the size of the training dataset.

For this Challenge, an augmented version of the Aff-Wild2 \cite{zafeiriou2017aff,kollias2017recognition,kollias2019expression,kollias2020analysing,kollias2021analysing,kollias2021affect,kollias2021distribution,kollias2022abaw,kollias2019face,kollias2023abaww} is used. This corpora is audiovisual (A/V), in-the-wild and in total consists of 594 videos of around 3M frames of 584 subjects.

The second Challenge is the  Expression (Expr) Recognition one; the target of this Challenge is to recognise between eight mutually exclusive classes in each frame of the utilized Challenge corpora; these classes are the 6 basic expressions (i.e., anger, disgust, fear, happiness, sadness and surprise), the neutral state and a category 'other' that denotes affective states that are not included in the neutral state or in the 6 basic expressions.

Only uni-task solutions will be accepted for this Challenge. Teams are allowed to use any -publicly or not- available pre-trained model (as long as it has not been pre-trained on the utilized in this Challenge corpora, Aff-Wild2). The pre-trained model can be pre-trained on any task (e.g., VA estimation, Expression Recognition, AU detection, Face Recognition). However when the teams are refining the model and developing the methodology, they should not use any other annotations (VA or AUs): the methodology should be purely uni-task, using only the Expr annotations. This means that teams are allowed to use other databases' Expr annotations, or generated/synthetic data (e.g. the data provided in the ECCV 2022 run of the ABAW Challenge \cite{kollias20222abaw}), or any affine transformations, or in general data augmentation techniques (e.g., \cite{psaroudakis2022mixaugment}) for increasing the size of the training dataset.

For this Challenge, the Aff-Wild2 is used. This corpora is audiovisual (A/V), in-the-wild and in total consists of 548 videos of around 2.7M frames.

The third Challenge is the  Action Unit (AU) Detection one; the target of this Challenge is to detect which of the 12 Action Units are activated in each frame of the utilized Challenge corpora. Action Units refer to a set of facial muscle movements or configurations. The action units that have been selected for the purposes of this Challenge are the following: AU1, AU2, AU4, AU6, AU7, AU10, AU12, AU15, AU23, AU24, AU25 and AU26.

Only uni-task solutions will be accepted for this Challenge. Teams are allowed to use any -publicly or not- available pre-trained model (as long as it has not been pre-trained on the utilized in this Challenge corpora, Aff-Wild2). The pre-trained model can be pre-trained on any task (e.g., VA estimation, Expression Classification, AU detection, Face Recognition). However when the teams are refining the model and developing the methodology, they should not use any other annotations (VA or Expr): the methodology should be purely uni-task, using only the AU annotations. This means that teams are allowed to use other databases' AU annotations, or generated/synthetic data, or any affine transformations, or in general data augmentation techniques (e.g., \cite{psaroudakis2022mixaugment}) for increasing the size of the training dataset.

For this Challenge, the Aff-Wild2 is used. This corpora is audiovisual (A/V), in-the-wild and in total consists of 542 videos of around 2.7M frames.

The fourth Challenge is the Compound Expression (CE)
Recognition one; the target of this Challenge is to recognise between the 7 mutually exclusive classes in each frame of the utilized Challenge corpora. These classes are the following compound expressions: Fearfully Surprised, Happily Surprised, Sadly Surprised, Disgustedly Surprised, Angrily Surprised, Sadly Fearful and Sadly Angry.

Teams are allowed to use any -publicly or not- available pre-trained model and any -publicly or not- available database (that contains any annotations, e.g. VA, basic or compound expressions, AUs).

For this Challenge, a part of C-EXPR-DB \cite{kollias2023multi} database will be used, which consists of 56 videos in total. C-EXPR-DB is an audiovisual (A/V) in-the-wild database and in total consists of 400 videos of around 200K frames. 

The final fourth challenge is the Emotional Mimicry Intensity (EMI) Estimation challenge where emotional mimics are explored. Participants are asked to predict six emotional dimensions using a multi-output regression approach. The following emotional expressions have been used: "Admiration", "Amusement", "Determination", "Empathic Pain", "Excitement", and "Joy".

For the purposes of this challenge, we use the audiovisual and in-the-wild HUME-Vidmimic2 dataset, a comprehensive collection derived from 'in-the-wild' settings, which contains more than 17\,000 videos totaling over 30 hours from the United States, similar to our first version~\cite{christ2023muse}.

The sixth ABAW Competition, which is part of the
respective Workshop to be held in conjunction with the IEEE Computer Vision and Pattern Recognition Conference (CVPR) 2024 is a continuation of the successful series of ABAW Competitions held in conjunction with IEEE CVPR 2023, ECCV 2022, IEEE CVPR 2022, ICCV 2021, IEEE FG 2020 and IEEE CVPR 2017, with the participation of many teams coming from both academia and industry, from all across the world \cite{zhang2023facial,zhang2023multimodal,wang2023spatio,savchenko2023emotieffnet,vu2023vision,wang2023facial,yin2023multi,zou2023spatial,kollias2023facernet,zhou2023continuous,zhang2023facial,savchenko2023emotieffnet,xue2023exploring,yu2023exploring,gera2023abaw,nguyen2023transformer,mutlu2023tempt,zou2023spatial,shu2023mutilmodal,kim2023multi,deng2020multitask,li2021technical,zhang2020m,do2020affective,chen2017multimodal,weichi,deng2021towards,zhang2021prior,vu2021multitask,wang2021multi,zhang2021audio,xie2021technical,jin2021multi,antoniadis2021audiovisual,oh2021causal,kuhnke2020two,gera2020affect,dresvyanskiy2020audio,youoku2020multi,liu2020emotion,gera2021affect,mao2021spatial,pahl2020multi,ji2020multi,han2016incremental,deng2020fau,saito2021action,meng2022multi,zhang2022continuous,nguyen2022ensemble,savchenko2022frame,karas2022continuous,rajasekar2022joint,zhang2022transformer,yu2022multi,kim2022facial,phan2022expression,xue2022coarse,jeong2022facial,wang2022facial,hoai2022attention,tallec2022multi,wang2022multi,jiang2022facial,deng2022multiple,haider2022ensemble,sun2022two,chang2022multi,gera2022ss,li2022facial,wang2022hybrid,nguyen2022multi,savchenko2022hse,li2022affective,zhang2022emotion,lee2022byel,jeong2022learning,lei2022mid,kollias2024distribution}.

\section{Competition Corpora}\label{corpora}

In the following, we present a brief synopsis of each Challenge's dataset. For a more comprehensive understanding, readers are encouraged to consult the original documentation. Additionally, we detail the pre-processing steps undertaken for the first three Challenges, which involve cropping and aligning all image-frames. These have been utilized in our baseline experiments. 

\subsection{Valence-Arousal Estimation Challenge}

\begin{figure}[h]
\centering
\includegraphics[height=7cm]{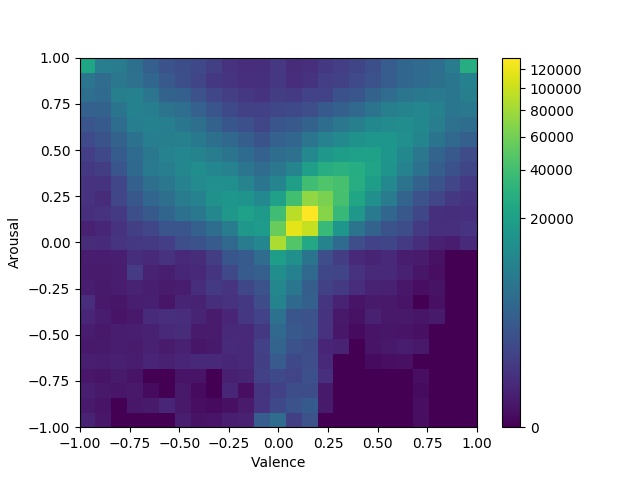}
\caption{Valence-Arousal Estimation Challenge: 2D Valence-Arousal Histogram of Annotations in Aff-Wild2}
\label{va_annot}
\end{figure}

This Challenge's dataset comprises $594$ videos, an expansion of the Aff-Wild2 database, annotated in terms of valence and arousal. Notably, sixteen videos feature two subjects, both of whom are annotated. In total, annotations are provided for 2,993,081 frames from $584$ subjects; these annotations have been conducted by four experts using the methodology outlined in \cite{cowie2000feeltrace}. Valence and arousal values are continuous and range in $[-1,1]$. The  2D Valence-Arousal histogram of annotations is visualized in Figure \ref{va_annot}.

Aff-Wild2 is split into training, validation and testing sets, in a subject independent manner, ensuring each subject appears exclusively in one set.  The training, validation and testing sets consist of 356, 76 and 162 videos, respectively.

The Train and Validation data along with their corresponding annotations are provided to the participating teams. The unlabeled test data are provided to the participating teams who will upload their test set predictions to an evaluation server. Participating teams are allowed to submit up to five sets of predictions for
this challenge.

\subsection{Expression Recognition Challenge}

In this Challenge, the dataset consists of 548 videos from Aff-Wild2, annotated for the six basic expressions, neutral state, and an 'other' category representing non-basic expressions. Seven videos feature two subjects, both of whom are annotated. In total, annotations are provided for  2,624,160 frames from 437 subjects. Annotation is conducted by seven experts on a frame-by-frame basis. Table \ref{expr_distr} presents the distribution of expression annotations.

\begin{table}[!h]
\caption{Expression Classification Challenge: Number of Annotated Images for each Expression  }
\label{expr_distr}
\centering
\begin{tabular}{ |c||c| }
\hline
 Expressions & No of Images \\
\hline
\hline
Neutral & 468,069  \\
 \hline
Anger & 36,627  \\
 \hline
Disgust & 24,412 \\
 \hline
Fear &  19,830 \\
 \hline
Happiness & 245,031  \\
 \hline
Sadness & 130,128  \\
 \hline
Surprise & 68,077  \\
 \hline
 Other & 512,262 \\
 \hline
\end{tabular}
\end{table} 
 
Aff-Wild2 is split into training, validation and testing sets, in a subject independent manner. 
Aff-Wild2 is split into training, validation and testing sets, in a subject independent manner. The training, validation and testing sets consist of 248, 70 and 230 videos, respectively.

The Train and Validation data along with their corresponding annotations are provided to the participating teams. The unlabeled test data are provided to the participating teams who will upload their test set predictions to an evaluation server. Participating teams are allowed to submit up to five sets of predictions for
this challenge.

\subsection{Action Unit Detection Challenge}

The dataset for this Challenge comprises 542 videos annotated for 12 AUs corresponding to the inner and outer brow raiser, the brow lowerer, the cheek raiser, the lid tightener, the upper lip raiser, the lip corner puller and depressor, the lip tightener and pressor, lips part and jaw drop. The exact utulized AUs along with their corresponding actions can be seen in Table \ref{au_distr}.
Annotations are provided for 2,627,632 frames from 438 subjects. A semi-automatic annotation procedure, involving both manual and automatic annotations, is employed. Table \ref{au_distr} further details the annotated AUs distribution.

\begin{table}[h]
    \centering
        \caption{Action Unit Detection Challenge: Distribution of AU Annotations in Aff-Wild2}
    \label{au_distr}
\begin{tabular}{|c|c|c|}
\hline
  Action Unit \# & Action   &\begin{tabular}{@{}c@{}} Total Number \\  of Activated AUs \end{tabular} \\   \hline    
    \hline    
   AU 1 & inner brow raiser   & 301,102 \\   \hline    
   AU 2 & outer brow raiser  & 139,936 \\   \hline   
   AU 4 & brow lowerer   & 386,689  \\  \hline    
   AU 6 & cheek raiser  & 619,775 \\  \hline    
   AU 7 & lid tightener  & 964,312 \\  \hline    
   AU 10 & upper lip raiser  & 854,519 \\  \hline    
   AU 12 & lip corner puller  & 602,835 \\  \hline    
   AU 15 & lip corner depressor  & 63,230 \\  \hline   
  AU 23 & lip tightener & 78,649 \\  \hline    
   AU 24 & lip pressor & 61,500 \\  \hline    
   AU 25 & lips part  & 1,596,055 \\  \hline     
   AU 26 & jaw drop  & 206,535 \\  \hline     
\end{tabular}
\end{table}

Aff-Wild2 is split into training, validation and testing sets, in a subject independent manner. The training, validation and testing sets consist of 295, 105 and 142 videos, respectively.

The Train and Validation data along with their corresponding annotations are provided to the participating teams. The unlabeled test data are provided to the participating teams who will upload their test set predictions to an evaluation server. Participating teams are allowed to submit up to five sets of predictions for
this challenge.

\subsection{Compound Expression Recognition Challenge}

For this Challenge, a part of C-EXPR-DB database will be used (56 videos in total). C-EXPR-DB is audiovisual (A/V) in-the-wild database and in total consists of 400 videos of around 200K frames; each frame is annotated in terms of 12 compound expressions. For this Challenge, the following 7 compound expressions will be considered: Fearfully Surprised, Happily Surprised, Sadly Surprised, Disgustedly Surprised, Angrily Surprised, Sadly Fearful and Sadly Angry.

Goal of the Challenge and Rules
Participants will be provided with a part of C-EXPR-DB database (56 videos in total), which will be unannotated, and will be required to develop their methodologies (supervised/self-supervised, domain adaptation, zero-/few-shot learning etc) for recognising the 7 compound expressions in this unannotated part, in a per-frame basis.

 \subsection{Emotional Mimicry Intensity Estimation Challenge}

In the Emotional Mimicry Intensity Challenge (EMI-Challenge), we investigate the study of emotional mimicry by presenting a large-scale and in-the-wild dataset featuring 557 participants and over 30 hours of audiovisual content. This dataset was collected in naturalistic settings, with participants using their webcams to record their facial and vocal responses by mimicking a "seed" video and rating it in the range from 0 to 100.

The data preparation process involved a speaker-independent partitioning of the dataset into training, validation, and test sets. Table~\ref{emi_partitions} statistics of the dataset for each partition. The train and validation data along with their corresponding annotations are provided to the participating teams. The unlabeled test data are provided to the participating teams who will upload their test set predictions to an evaluation server.

Along with the data, the participants are provided the faces of individuals within the videos that were detected with the use of MTCNN~\cite{zhang2016joint} at a frequency of 6 frames per second. In addition, features extracted from the raw signals and thus enabling participants to use end-to-end approaches~\cite{tzirakis2017end, tzirakis2021end, tzirakis2018end2you, tzirakis2021speech, tzirakis2018end} are provided. Specifically, the feature sets provided are the Vision Transformer (ViT)~\cite{caron2021emerging} for the faces and Wav2Vec 2.0~\cite{baevski2020wav2vec} for the audio signals.

\begin{table}[h!]
\centering
\begin{tabular}{|l|r|r|}
\hline
Partition & Duration & \# Samples \\ 
 & (HH:MM:SS) & \\ \hline \hline
Train & 15:07:03 & 8072 \\ \hline
Validation & 9:12:02 & 4588 \\ \hline
Test & 9:04:05 & 4586 \\ \hline
\end{tabular}
\caption{HUME-Vidmimic2 partition statistics.}
\label{emi_partitions}
\end{table}

\subsection{Aff-Wild2 Pre-Processing: Cropped \& Cropped-Aligned Images} \label{pre-process}


Initially, all videos are segmented into individual frames, after which they undergo processing using the RetinaFace detector. This step aims to extract face bounding boxes and five facial landmarks for each frame. Subsequently, the images are cropped based on the bounding box coordinates, and these cropped images are provided to the participating teams.

Using the five facial landmarks (representing two eyes, the nose, and two mouth corners), a similarity transformation is applied. This transformation ensures alignment, resulting in cropped and aligned images, which are also shared with the participating teams. Ultimately, these cropped and aligned images are utilized in our baseline experiments.

All cropped and cropped-aligned images are resized to dimensions of $112 \times 112 \times 3$ pixels and their intensity values are normalized to fall within the range of $[-1, 1]$.

\section{Evaluation Metrics Per Challenge}\label{metrics}


\subsection{Valence-Arousal Estimation Challenge}

The performance measure is the average between the Concordance Correlation Coefficient (CCC) of valence and arousal:

\begin{equation} \label{va}
\mathcal{P}_{VA} = \frac{CCC_a + CCC_v}{2}
\end{equation}

CCC evaluates the agreement between two time series (e.g., all video annotations and predictions) by scaling their correlation coefficient with their mean square difference. In this way, predictions that are well correlated with the annotations but shifted in value are penalized in proportion to the deviation. CCC takes values in the range $[-1,1]$, where $+1$ indicates perfect concordance and $-1$ denotes perfect discordance. The highest the value of the CCC the better the fit between annotations and predictions, and therefore high values are desired.
CCC is defined as follows:

\begin{equation} \label{ccc}
CCC  = \frac{2s_x  s_y \rho_{xy}}{s_x^2 + s_y^2 + (\bar{x} - \bar{y})^2},
\end{equation}

\noindent
where $\rho_{xy}$ is the Pearson's Correlation Coefficient, $s_x$ and $s_y$ are the variances of all video valence/arousal annotations and predicted values, respectively and $s_{xy}$ is the corresponding covariance value.


\subsection{Expression Classification Challenge}\label{evaluation}

The performance measure is the average F1 Score across all 8 categories (i.e., macro F1 Score):

\begin{equation} \label{expr}
\mathcal{P}_{EXPR} = \frac{\sum_{expr} F_1^{expr}}{8}
\end{equation}

The $F_1$ score is a weighted average of the recall (i.e., the ability of the classifier to find all the positive samples) and precision (i.e., the ability of the classifier not to label as positive a sample that is negative). The $F_1$ score  takes values in the range $[0,1]$; high values are desired. The $F_1$ score is defined as:

\begin{equation} \label{f1}
F_1 = \frac{2 \times precision \times recall}{precision + recall}
\end{equation}


\subsection{Action Unit Detection Challenge}\label{evaluation3}

The performance measure is the average F1 Score across all 12 AUs. Therefore, the evaluation criterion for the  Action Unit Detection Challenge is:

\begin{equation} \label{au}
\mathcal{P}_{AU} = \frac{\sum_{au} F_1^{au}}{12}
\end{equation}

\subsection{Compound Expression Recognition Challenge}\label{evaluation4}

The performance measure is the average F1 Score across all 7 compound expressions. Therefore, the evaluation criterion for the  Compound Expression Recognition Challenge is:

\begin{equation} \label{ce}
\mathcal{P}_{CE} = \frac{\sum_{expr} F_1^{expr}}{7}
\end{equation}

\subsection{Emotional Mimicry Intensity Estimation Challenge}\label{mtl}

The performance measure is the average Pearson’s Correlation Coefficient ($\rho$) across the 6 emotion dimensions:

\begin{equation} \label{ppp}
\mathcal{P}_{EMI} = \frac{\sum_{i=1}^{6} \rho^{i}}{6}
\end{equation}

Pearson’s Correlation Coefficient ($\rho$) takes values in the range $[-1,1]$; high values are desired.

\section{Participating Teams' and Baseline Methods' Results} \label{baseline}

All baseline systems are built solely on existing open-source machine learning toolkits to maintain result reproducibility. TensorFlow is the chosen framework for implementing all systems.

In this Section, we describe the baseline systems developed for each Challenge and present their  obtained results.

\subsection{Valence-Arousal Estimation Challenge}

The baseline network comprises a ResNet architecture with 50 layers, initially trained on ImageNet (ResNet50). It incorporates a linear output layer responsible for providing the final estimations for valence and arousal. Its average CCC performance on the validation set is 0.22 (the CCC for valence is 0.24 and the CCC for arousal is 0.20), as can be seen in Table \ref{comparison_sota_va}.

\begin{table}[ht]
\caption{Valence-Arousal Estimation Challenge's Results; Total Score is the average CCC between valence and arousal} 
\label{comparison_sota_va}
\centering
\begin{tabular}{ |c||c||c|c| }
 \hline
\multicolumn{1}{|c||}{\begin{tabular}{@{}c@{}} Teams \end{tabular}}  & \multicolumn{1}{c||}{\begin{tabular}{@{}c@{}}Total Score \end{tabular}} &
\multicolumn{1}{c|}{\begin{tabular}{@{}c@{}} CCC-V  \end{tabular}} &
\multicolumn{1}{c|}{\begin{tabular}{@{}c@{}}  CCC-A \end{tabular}}
\\ 
  \hline
 \hline

baseline & 0.22
&  \begin{tabular}{@{}c@{}} 0.24  \end{tabular} 
& \begin{tabular}{@{}c@{}}  0.20   \end{tabular}
 \\
\hline

\end{tabular}
\end{table}

\subsection{Expression Classification Challenge}

The baseline network adopts a VGG16 architecture with fixed convolutional weights (i.e., non-trainable), while only the three fully connected layers are trainable. It is pre-trained on the VGGFACE dataset and equipped with an output layer featuring a softmax activation function, facilitating the prediction of eight expressions. Mixaugment has been used as data augmentation technique. 

MixAugment is a simple and data-agnostic data augmentation routine that trains a method on convex combinations of pairs of examples and their labels. It extends the training distribution by incorporating the prior knowledge that linear interpolations of feature vectors should lead to linear interpolations of the associated targets. 
MixAugment constructs virtual training examples $(\tilde{x},\tilde{y})$ as follows:
\begin{align}
\tilde{x} &= \lambda x_i+(1-\lambda)x_j  \nonumber \\ 
\tilde{y} &= \lambda y_i+(1-\lambda)y_j
\label{eq:mixup}
\end{align}

\noindent where $x_i$ and $x_j$ are two random raw inputs (i.e., images), $y_i$ and $y_j$ $\in \{0,1\}^8$ are their corresponding one-hot label encodings  and  $\lambda \thicksim \mathrm{B}(\alpha,\alpha) \in [0,1]$ (i.e., Beta distribution) for $\alpha \in (0, \infty)$. 

During each training iteration, the baseline network is trained concurrently on both real (r) and virtual (v) examples. Specifically, in each training iteration, the network is fed with both $x_i$ and $x_j$, and the generated image $\tilde{x}$ (of Eq. \ref{eq:mixup}).

The  performance of the baseline network on the validation set is 0.25, as can be seen in Table \ref{comparison_sota_expr}.

\begin{table}[ht]
\caption{Expression Recognition Challenge's Results;  the evaluation criterion is the average F1 Score} 
\label{comparison_sota_expr}
\centering
\begin{tabular}{ |c||c| }
 \hline
\multicolumn{1}{|c||}{\begin{tabular}{@{}c@{}} Teams \end{tabular}}  & \multicolumn{1}{c|}{\begin{tabular}{@{}c@{}}Total Score \end{tabular}}
\\ 
  \hline
 \hline

baseline without MixAugment & 0.23
 \\
baseline with MixAugment & 0.25
 \\
 
\hline

\end{tabular}
\end{table}

\subsection{Action Unit Detection Challenge}

The baseline network adopts a VGG16 architecture with fixed convolutional weights (i.e., non-trainable), while only the three fully connected layers are trainable. It is pre-trained on the VGGFACE dataset and equipped with an output layer featuring a sigmoid activation function, facilitating the detection of the twelve AUs.  Its  performance on the validation set is 0.39, as can be seen in Table \ref{comparison_sota_au}.

\begin{table}[ht]
\caption{Action Unit Detection Challenge's Results;  the evaluation criterion is the average F1 Score} 
\label{comparison_sota_au}
\centering
\begin{tabular}{ |c||c| }
 \hline
\multicolumn{1}{|c||}{\begin{tabular}{@{}c@{}} Teams \end{tabular}}  & \multicolumn{1}{c|}{\begin{tabular}{@{}c@{}}Total Score \end{tabular}}
\\ 
  \hline
 \hline

baseline & 0.39
 \\
\hline

\end{tabular}
\end{table}

\subsection{Compound Expression Recognition Challenge}

No baseline network and results will be provided for this Challenge due to the nature of the Challenge.

\subsection{Emotional Mimicry Intensity Estimation Challenge}

We set initial baseline results with two distinct sets of features. Initially, we utilized features derived from a pre-trained Vision Transformer (ViT), which were then processed by a three-layer Gated Recurrent Unit (GRU) network, achieving a performance score of $\mathcal{P}_{EMI}=0.09$. Subsequently, we leveraged features extracted from Wav2Vec2, paired with a linear processing layer, attaining a performance score of $\mathcal{P}_{EMI}=0.24$. Furthermore, by averaging the predictions from both of these unimodal techniques, we pursued a multimodal strategy that culminated in an enhanced combined performance score of $\mathcal{P}_{EMI}=0.25$.

\begin{table}[ht]
\caption{Baseline results on the validation set for the EMI challenge.} 
\label{comparison_sota_au}
\centering
\begin{tabular}{ |c||c| }
 \hline
\multicolumn{1}{|c||}{\begin{tabular}{@{}c@{}} Model \end{tabular}}  & \multicolumn{1}{c|}{\begin{tabular}{@{}c@{}}Total Score \end{tabular}}
\\  \hline \hline
Audio & 0.24 \\
Vision & 0.09 \\
baseline & 0.25 \\
\hline
\end{tabular}
\end{table}

\section{Conclusion}\label{conclusion}

In this paper we have presented the sixth Affective Behavior Analysis in-the-wild Competition (ABAW)  held in conjunction with IEEE CVPR 2024. This Competition is a continuation of the series of ABAW Competitions.  This Competition comprises five Challenges targeting: i)  Valence-Arousal Estimation, ii)  Expression Recognition (8 categories), iii)  Action Unit Detection (12 action units), iv) Compound Expression Recognition (7 categories)  and v) Emotional Mimicry Intensity Estimation (6 emotion dimensions). The databases utilized for this Competition are an extended version of Aff-Wild2, the C-EXPR-DB and the Hume-Vidmimic2 dataset.

{
    \small
    \bibliographystyle{ieeenat_fullname}
    \bibliography{main}
}


\end{document}